\documentclass[10pt,twocolumn,letterpaper]{article}

\usepackage{wacv}
\usepackage{times}
\usepackage{epsfig}
\usepackage{graphicx}
\usepackage{amsmath}
\usepackage{amssymb}
\usepackage{times}
\usepackage{multirow} 
\usepackage{diagbox, eqparbox, hhline}
\usepackage{algorithmicx}

\usepackage{mathrsfs,subcaption}
\usepackage{algpseudocode}

\usepackage[dvipsnames]{xcolor}
\usepackage[ruled,vlined,linesnumbered]{algorithm2e}

\wacvfinalcopy 

\def\wacvPaperID{1094} 

\ifwacvfinal\fi
\begin{document}

\title{Robust Facial Landmark Detection via Aggregation on Geometrically Manipulated Faces }
\author{Seyed Mehdi Iranmanesh, Ali Dabouei, Sobhan Soleymani, Hadi Kazemi, Nasser M. Nasrabadi \\
West Virginia University\\
{\tt\small \{seiranmanesh,ad0046,ssoleyma,hakazemi\}@mix.wvu.edu, \{nasser.nasrabadi\}@mail.wvu.edu}}

\maketitle
\ifwacvfinal\thispagestyle{empty}\fi

\begin{abstract}
  In this work, we present a practical approach to the problem of facial landmark detection. The proposed method can deal with large shape and appearance variations under the rich shape deformation. To handle the shape variations we equip our method with the aggregation of manipulated face images. The proposed framework generates different manipulated faces using only one given face image. The approach utilizes the fact that small but carefully crafted geometric manipulation in the input domain can fool deep face recognition models.
  We propose three different approaches to generate manipulated faces in which two of them perform the manipulations via adversarial attacks and the other one uses known transformations. Aggregating the manipulated faces provides a more robust landmark detection approach which is able to capture more important deformations and variations of the face shapes. Our approach is demonstrated its superiority compared to the state-of-the-art method on benchmark datasets AFLW, 300-W, and COFW.           
\end{abstract}


\section{Introduction}
Facial landmark detection goal is to identify the location of predefined facial landmarks (\textit{i.e.}, tip of the nose, corner of the eyes, and eyebrows). Reliable landmark estimation is part of the procedure for more complicated vision tasks. It can be applied to the variant tasks such as 3D face reconstruction~\cite{liu2016joint}, head pose estimation~\cite{wu2017simultaneous}, facial reenactment~\cite{thies2016face2face}, and face recognition~\cite{zhu2015high}. However, it remains challenging due to the necessity of handling non-rigid shape deformations, occlusions, and appearance variations. For example, facial landmark detection must handle not only coarse variations such as illumination and head pose but also finer variations including skin tones and expressions. 
\begin{figure}[t]
\begin{center}
 \includegraphics[width=0.7\linewidth]{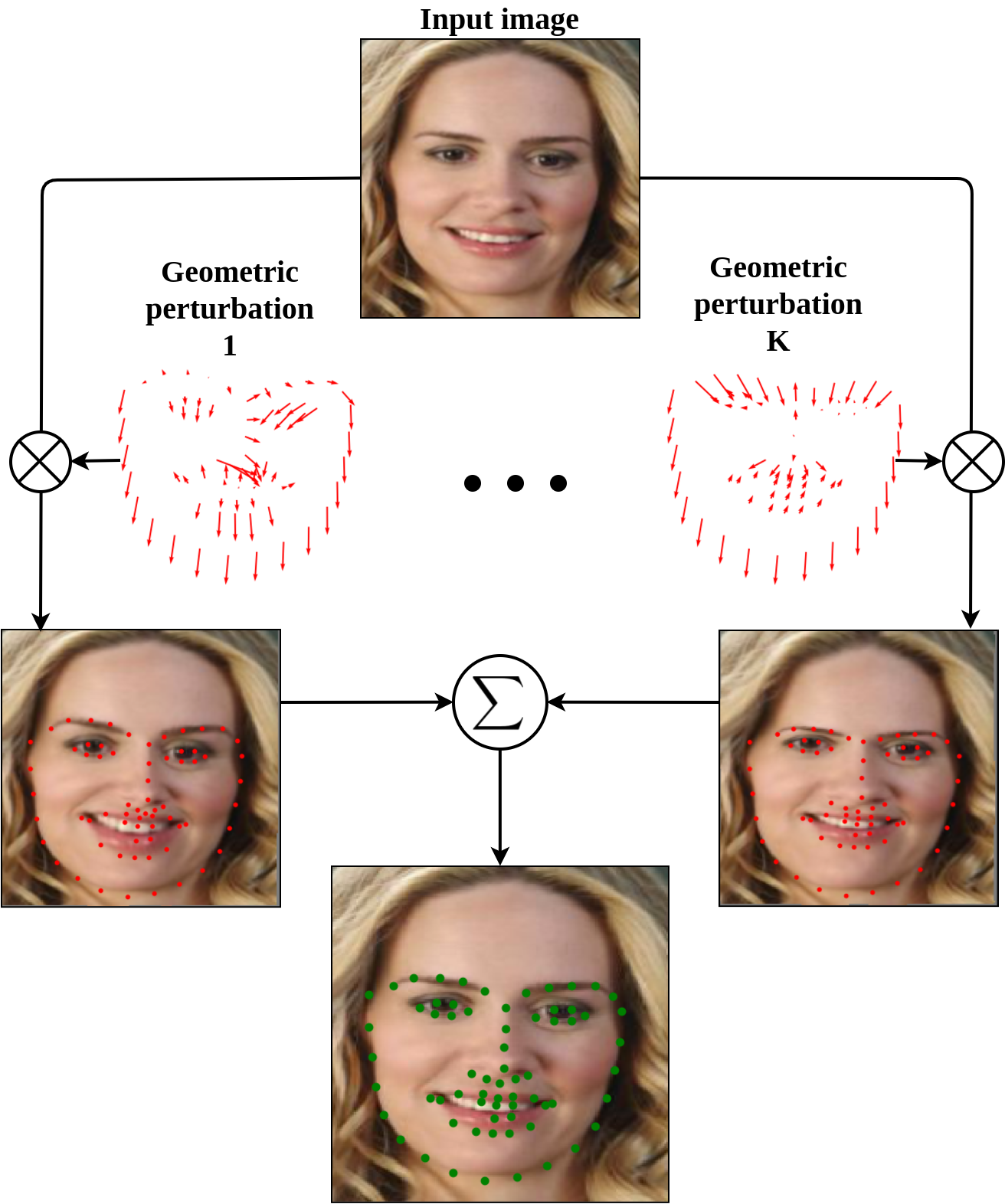}
   
\end{center}
   \caption{An input face image is manipulated utilizing geometric perturbations that target important locations of face images for the task of landmark detection. $K$ different manipulated faces are generated where each of them contains the important displacements from the input image. The aggregation on these manipulated images leads to robust landmark detection.}
\label{fig:long}
\label{fig:onecol}
\end{figure}
There has been a wide range of approaches to solve the problem of landmark detection, starting with methods such as active shape model~\cite{cootes1995active}, and active appearance models~\cite{cootes2001active} which are related to PCA-based shape constraint. 

Many of these approaches utilize a cascade strategy to integrate prediction modules and update the landmark locations in a progressive manner~\cite{zhang2014coarse,dong2017more}. Cascade regression networks which are designed for landmark localization~\cite{sun2013deep}, or human body pose estimation~\cite{toshev2014deeppose} have made improvements by tackling the problem level at coarse to fine levels. However, requiring careful design and initialization for such frameworks and the absence of learned geometric relationships are the main challenges of these architectures.      

Recently, with the onset of convolutional neural networks (ConvNets) in feature representation~\cite{simonyan2014very},  a common approach in facial landmark detection is to extract features from the facial appearance using ConvNets, and afterward learn a model typically a regressor to map the features to the landmark locations~\cite{zhang2014coarse,dollar2010cascaded,peng2016recurrent,ren2014face}. Despite the excellent performance of the ConvNets in different applications, it has been shown~\cite{goodfellow2014explaining,szegedy2013intriguing} that they can be very sensitive and vulnerable to a small perturbation in the input  domain  which  can  lead to a drastic change of the output domain, \textit{e.g.}, predicted landmarks.

Many approaches solve the face alignment problem with multi-tasking approaches. However, the task of face alignment might not be in parallel with the other tasks. For example, in the classification task, the output needs to be invariant to small deformations such as translation. However, in tasks such as landmark localization or image segmentation both the global integration of information as well as maintaining the local information and pixel-level detail is necessary. The goal of precise landmark localization has led to evolving new architectures such as dilated convolutions~\cite{yu2015multi}, recombinator-networks~\cite{honari2016recombinator}, stacked what where auto-encoders~\cite{zhao2015stacked}, and hyper-columns~\cite{hariharan2015hypercolumns} where each of them attempts to preserve pixel-level information.         

In this paper, we propose a geometry aggregated network (GEAN) for face alignment which can comfortably deal with rich expressions and arbitrary shape variations. We design a novel aggregation framework which optimizes the landmark locations directly using only one image without requiring any extra prior which leads to robust alignment given arbitrary face deformations. We provide three different approaches to produce deformed images using only one image and aggregate them in a weighted manner according to their amount of displacement to estimate the final locations of the landmarks. Extensive empirical results indicate the superiority of the proposed method compared to existing methods on challenging datasets with large shape and appearance variations, \textit{i.e.}, 300-W~\cite{sagonas2013300} and ALFW~\cite{6130513}.

\section{Related Work} 

A common approach to facial landmark detection problem is to leverage deep features from ConvNets. These facial features and regressors are trained in an end-to-end manner utilizing a cascade strategy to update the landmark locations progressively~\cite{sun2013deep, 7780740}. Yu et al.~\cite{DeepDeformation} integrate geometric constraints within CNN architecture using a deep deformation network. Lev et al.~\cite{8099876} propose a deep regression framework with two-step re-initialization to avoid the initialization issue. Zhu et al.~\cite{7299134} also tried to deal with poor initialization utilizing a coarse search over a shape space with variant shapes. In another work, Zhu et al.~\cite{7780740}, overcome the extreme head poses and rich shape deformations exploiting cascaded regressors.

Another category of landmark detection approaches leverages the end-to-end training from ConvNets frameworks to learn robust heatmaps for landmark detection task~\cite{7780880,8237378,stacked}. Balut et al.~\cite{8237378} utilized the residual framework to propose a robust network for facial landmark detection.  Newell et al.~\cite{stacked} and Wei et al.~\cite{7780880} consider the coordinate of the highest response on the heatmaps as the location of landmarks for human pose estimation task.  

In a more general definition, this problem can also be viewed as learning structural representation. Some studies~\cite{Reed:2014,NIPS2015_5845}, disentangle visual content into different factors of variations such as camera viewpoint, motion and identity to capture the inherent structure of objects. However, the physical parameters of these factors are embedded in a latent representation which is not discernible. Some methods can handle~\cite{7553523,8237584} conceptualize structures in the multi-tasking framework as auxiliary information (\textit{e.g.}, landmarks, depth, and mask). Such structures in these frameworks are designed by humans and need supervision to learn.

\section{Proposed Method}

Given a face image $I \in \mathbb{R}^{w\times h}$ with spatial size $W\!\times\!H$, the facial landmark detection algorithm aims to find a prediction function $\Phi: \mathbb{R}^{W\times H} \rightarrow \mathbb{R}^{2 \times L}$ which estimates the 2D locations of $L$ landmarks. We seek to find a robust and accurate version of $\Phi$ by training a deep function through the aggregation of geometrically manipulated faces. The proposed method consists of different parts which will be described in detail. 

\subsection{Aggregated Landmark Detector.}

The proposed approach attempts to provide a robust landmark detection algorithm to compensate for the lack of a specific mechanism to handle arbitrary shape variations in the literature of landmark detection. The method builds upon aggregating set of manipulated images to capture robust landmark representation.
Given a face image $I$, a set of manipulated images are constructed such that $\hat{I}_k = M (I,\theta_k)$ is the $k$-th manipulated face image and $\theta_k$ is its related parameters for the manipulating function $M$. Considering the set of manipulated images, we seek a proper choice of $M$ such that aggregating landmark information in the set $\{\Phi (\hat{I}): k=1 \dots K\}$ provides a more accurate and robust landmark features compared to $\Phi(I)$ which solely uses the original image $I$. Therefore, one important key in the aggregated method is answering the question of ``how'' to manipulate images. Face images typically have a semantic structure which have a similar global structure but the local and relative characteristics of facial regions differ between individuals. Hence, a straightforward and comprehensive choice of the manipulation function $M$ should incorporate the prior information provided by the global consistency of semantic regions and uniqueness of relative features which can be interpreted as the ID information. Hence, we build our work based on a choice of $M$ which incorporates geometric transformations to manipulate relative characteristics of inputs samples while preserving the semantic and global structure of input faces.    

To incorporate ID information, we consider a pretrained face recognizer $f: \mathbb{R}^{W\!\times \!H}\rightarrow \mathbb{R}^{n_z}$ mapping an input face image to an ID representation $z \in \mathbb{R}^{n_z}$, where cardinality of the embedding subspace is $n_z$ (typically set to be $128$~\cite{schroff2015facenet}). Having $f$ makes it possible to compare IDs of two samples by simply measuring the $\ell_2$-norm of their representation in the embedding space. Hence, we geometrically manipulate the input face image to change its ID. It should be noted that since $f$ is trained on face images, the corresponding embedding space of IDs captures a meaningful representation of faces. Therefore, the manipulated faces contain rich information with regards to face IDs. 

 
 
To manipulate the face image $I$ based on landmark coordinates, we consider coarse landmark locations $P = \{(x_0,y_0),\dots, (x_{L-1},y_{L-1})\}$ and define the displacement field $d$ to manipulate the landmark locations. Given the $i$-th source landmark $(x_i, y_i)$, we compute its manipulated version using the displacement vector $d_i = (\Delta x_i, \Delta y_i)$. The manipulated landmark $p_i+d_i$ is as follows: 
\begin{equation}
p_i+d_i = (x_i+ \Delta x_i,y_i+ \Delta y_i)  \;.
\label{eq-adver}
\end{equation}
\begin{figure*}[t]
\begin{center}

\includegraphics[width=0.8\linewidth]{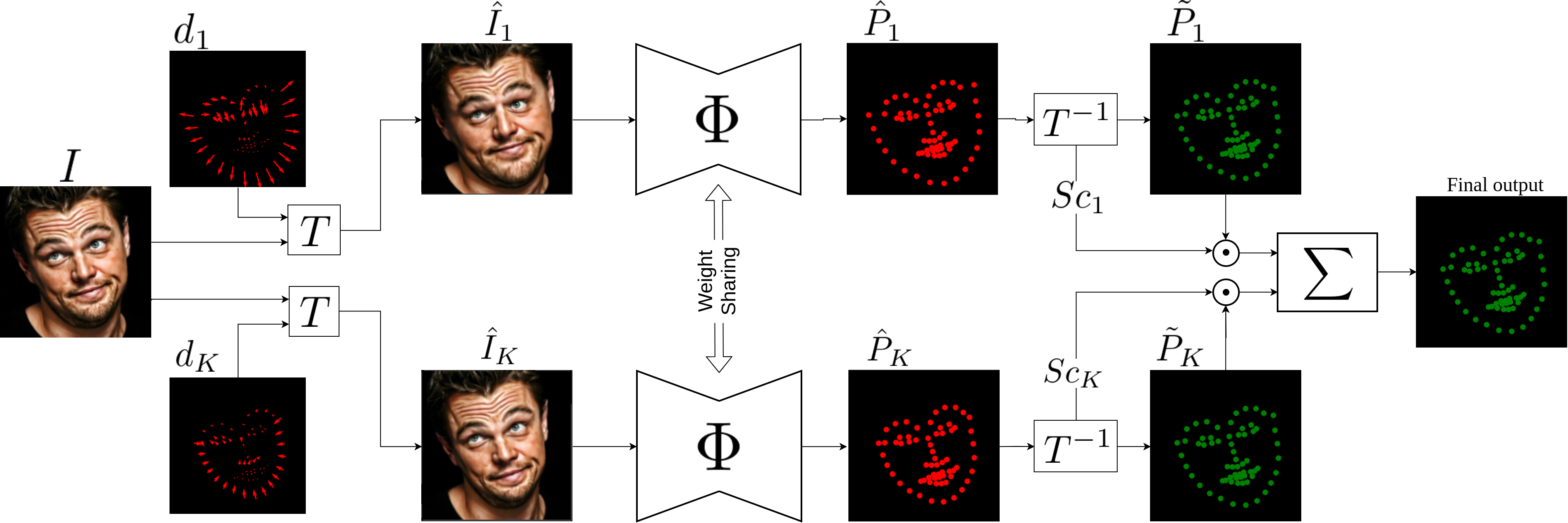}
\end{center}
   \caption{The overview of the proposed aggregated framework (GEAN). It consists of four steps: 1) $K$ different manipulated faces are generated; 2) Each manipulated face is given to the shared landmark detector $\Phi$ to extract its landmarks; 3) The inverse of transformation matrix is applied to the extracted landmarks to compensate for the displacement of step 1; 4) The normalization score values for each landmark of each branch is calculated and the aggregation is performed to extract the final landmark locations.}

\label{fig:land_diagram}
\end{figure*}
We present three different approaches to find a proper displacement ($d$) for manipulating face images. 
\subsection{Manipulation by Adversarial Attack.} \label{Landmarkadv} In the first approach we use adversarial attacks~\cite{43405} to manipulate facial landmarks to fool a face recognizer. Xiao et al.~\cite{Xiao2018SpatiallyTA}, proposed stAdv attack to generate adversarial examples using spatially transforming benign images. They utilize a displacement field for all the pixels in the input image. Afterward, they computed the corresponding location of pixels in the adversarial image using the displacement field $d$. However, optimizing a displacement field for all the pixels in the image is a highly non-convex function. Therefore, they used the L-BFGS~\cite{Liu1989}, with a linear backtrack search to find the optimal displacement field which is computationally expensive. Here, our approach considers the fact that the facial landmarks provide highly discriminative information for face recognition tasks~\cite{1333734}. In fact, face recognition tasks are highly linear around the original coordinates of the facial landmarks as it is shown in~\cite{Dabouei2019FastGA}.

In contrast to~\cite{Xiao2018SpatiallyTA} which computes the displacement field for all the pixels, our proposed method is inspired by~\cite{Dabouei2019FastGA} and estimates the $d$ only for $L$ landmarks and it does not suffer from the computational complexity. In addition, it is possible to apply the conventional spatial transformation to transform image. Therefore, the adversarial (manipulated) image using the transformation $T$  is as follows:

\begin{equation}
\hat{I} = T (P, P+d, I) \;,
\label{eq-adver}
\end{equation}
 
\noindent  where $T$ is the thin plate spline (TPS)~\cite{24792} transformation mapping from the source landmarks (control points) $P$ to the target ones $P+d$. In order to make the whole framework differentiable with respect to the landmark locations, we select a  differentiable interpolation function (\textit{i.e.}, differentiable bilinear interpolation)~\cite{NIPS2015_5854} so that the prediction of the face recognizer is differentiable with respect to the landmark locations. 

In this approach, we employ the gradient of the prediction in a face recognition model to update the displacement field $d$ and geometrically manipulate the input face image. We extend Dabouei et al. ~\cite{Dabouei2019FastGA} work in a way to generate $K$ different adversarial faces where each face represents a different ID ($K$ different IDs will be generated). Considering an input image $I$,  a face recognizer $f$ , and a set of $k-1$ manipulated images $S_I = \{\hat{I}_1, ...., \hat{I}_{k-1}\}$ the cost is defined as follows for the $k$-th adversarial face:

\begin{equation}
\mathcal{L} = \sum\limits_{I' \in S_{I}} ||f(T(P,P+d,I))-f(I')||_2\;.
\label{eq-adver}
\end{equation}

Inspired by FGSM~\cite{43405}, we employ the direction of the gradients of the prediction to update the adversarial landmark locations $P+d$, in an iterative manner. Considering $P+d$ as $P^{adv}$, using FGSM~\cite{43405}, the $t$-th step of optimization is as follows:
\begin{equation}
P^{adv}_t = P^{adv}_{t-1} + \epsilon  \; sign (\nabla_{P^{adv}_{t-1}} \mathcal{L}) \;.
\label{eq-4}
\end{equation}

In addition, we consider the clipping technique to constrain the displacement field in order to prevent the model from generating distorted face images. The algorithm continues the optimization for the $k$-th landmark locations until $\underset{I'\in S_I}{min} \{||f(\hat{I})-f(I')||_2 \} < \tau$ is failed, where $\tau$ is simply the distance threshold in the embedding space. In this way, we make sure that the $k$-th manipulated face has a minimum distance of $\tau$ to the other manipulated images in the face embedding subspace. Algorithm~\ref{alg1} shows the proposed procedure for generating $K$ different manipulated faces.   




\subsection{Manipulation of Semantic Groups of Landmarks using Adversarial Attacks.}\label{Landmark_Groupadv} In the first approach, we consider a fast and efficient approach to generate different faces based on the given face image. However, the first approach does not directly consider the fact that different landmarks semantically placed in different groups (\textit{i.e.}, landmarks related to lip, left eye, right eye, etc.). This might lead to generating severely distorted adversarial images. 

We added the clipping constraint to mitigate this issue in the first approach. Here, we perform semantic landmarks grouping \cite{Dabouei2019FastGA}. We categorize the landmarks into $n$ semantic groups $P_i, i \in \{1,\dots n\}$, where $p_{i,j}$ denotes the $j$-th landmark in the group $i$ which contains $c_i$ landmarks. These groups are formed based on different semantic regions which construct the face shape (\textit{i.e.}, lip, left eye, right eye, etc.). Semantic landmark grouping considers a scale and translation for each semantic group, instead of independently displacing each landmark. This consideration allows us to increase the total amount of displacement while preserving the global structure of the face.

\begin{algorithm}[t]
\small
\textbf{Input:} Image $I$, number of branches $K$, face recognizer $f$, distance threshold $\tau$, clipping threshold $\delta$.

\textbf{Output:} Set of adversarial faces $S = \{\hat{I}_1,..., \hat{I}_K\}$.

Initialize $\hat{I} \gets I$ and $S = \{I\}$.

\For {$k=1$ \KwTo $K$}{

$\hat{I}_{t=0,k} \gets I$;

\While{$\underset{I'\in S_I}{min} \{||f(\hat{I}_{t,k})-f(I')||_2 \} < \tau$}{
$\mathcal{L}= \sum\limits_{I' \in S_{I}} ||f(T(P,P^{adv}_{t},I))-f(I')||_2$;

$P^{adv}_{t+1} = P^{adv}_{t} + \epsilon  \; sign (\nabla_{P^{adv}_{t}} \mathcal{L})$;

$P^{adv}_{t+1} = clip \; (P^{adv}_{t+1}, \delta) $;

$\hat{I}_{t+1,k} = T (P, P_{t+1}^{adv},I)$;

}

$S \gets \{S, \hat{I}_k\}$;
}
return $S- \{I\}$;
\caption{Adversarial Face Generation}
\label{alg1}
\end{algorithm}

Let $P_i$ represents the $i$-th landmark group (\textit{e.g.}, group of landmarks related to the lip). The adversarial landmark locations which are semantically grouped can be obtained as following:
\begin{equation}
P_i^{adv} = \alpha_i (P_i - \bar{p_i}) + \beta_i\;,
\label{eq-adver}
\end{equation}
\noindent where $\bar{p_i}= \dfrac{1}{c_i}\sum_{j=1}^{c_i} p_{i,j}$ is the average location of all the landmarks in group $P_i$, and $\alpha_i$ and $\beta_i$ for each group can be computed using the closed-from solution in ~\cite{Dabouei2019FastGA}. It should be noted that the value of displacement $d$ for each branch used for computing semantic scales and translations is obtained using Algorithm \ref{alg1}. The only difference is that we add semantic perturbations constructed by Eq. \ref{eq-adver} instead of the random perturbation in line 8 of Algorithm \ref{alg1}. Therefore, the scale and translation of each semantic part of the face is different from other manipulated images in the set $S$.   


\subsection{Manipulation of Semantic Group of Landmarks with Known Transformation.}\label{Landmark_Groupknown}

In this approach, we semantically group the landmark locations in the same manner as the previous approach. Afterward, we uniformly sample ranges $[0.9, 1.1]^2$ and $[-0.05\!\times\!W, 0.05\!\times\!W]^2$ for the scale and translation of each semantic group, respectively. It may be noted that in the post-processing stage, we make sure that the semantic inter-group structure is preserved, \eg, eyes region does not interfere with the eyebrows region and they are symmetric according to each other. Therefore, the heuristic post-processing limits the above ranges based on the properties of each group. For instance, eyebrows could achieve higher vertical displacement compared to eyes since there is no semantic part above them to impose a constraint.


\subsection{Landmark Detector.}

Next, the hourglass network proposed in~\cite{stacked} is employed to estimate the facial landmarks location. Hourglass is designed based on residual blocks~\cite{He2016DeepRL}. It is a symmetric top-down and bottom-up fully convolutional network. The residual modules are able to capture high-level features based on the convolutional operation, while they can maintain the original information with the skip connections. The original information is branched out before downsampling and concatenated together before each up-sampling to retain the resolution information. Therefore, hourglass is an appropriate topology to capture and consolidate information from different resolutions and scales. 


After manipulating the face images (employing either of the three aforementioned approaches), we employ the hourglass network, $\Phi$, to extract the landmarks from the manipulated images. The network $\Phi$ is shared among all the branches of the framework as it is shown in Fig.~\ref{fig:land_diagram}.  Each landmark has a corresponding detector, which convolutionally extracts a response map. Taking $r_i$ as a $i$-th response map, we use the weighted mean coordinate as the location of the $i$-th landmark as follows:  

\begin{equation}
\hat{p}_i = (x_i,y_i) = \dfrac{1}{\zeta_i} \sum\limits_{u=1}^H \sum\limits_{v=1}^W (u,v).r_i(u,v) \;,
\label{eq-hourg}
\end{equation}
\noindent where $H$ and $W$ are the height and width of the response map which are the same as the spatial size of the input image, and $\zeta_i = \sum_{u, v} r_i(u, v)$. 

\subsection{Aggregation.}
After extracting the facial landmarks using the shared landmark detector $\Phi$ for each of the manipulated face images, we aim to move the predicted landmarks $\hat{P}$ toward their original locations. Let $T$ be a transformation that is used to convert the original faces to the manipulated ones (\textit{i.e.}, via adversarial attack approaches or the known transformation approach). We employ the inverse of the transformation matrix on the predicted landmarks to compensate for the displacement of them and denote the new landmark locations as $\tilde{P}$.         

The proposed approach contains a set of landmarks from $K$ branches, \textit{i.e.}, $\tilde{P} = \{\tilde{p}_{i,k}\}$ in which $i \in \{1, \dots, L\}$, and $k \in  \{1,\dots,K\}$ is the $i$-th landmark location in $k$-th branch of the framework. Each branch considers a score value which normalizes the displacement of landmarks caused by the manipulation approach (\textit{i.e.}, via adversarial attacks or known transformations) in each branch of the aggregated network as follows:

\begin{equation}
 Sc_{i,k} =\dfrac{\sqrt{\Delta x_{i,k} ^2 +  \Delta y_{i,k} ^2 }}{\sum\limits_{k=1} ^K \sqrt{\Delta x_{i,k} ^2 +  \Delta y_{i,k} ^2 }}    \;,
\label{normalization}
\end{equation}

\noindent where $Sc_{i,k}$ represents the displacement value for the $i$-landmark in the $k$-th branch. This score is utilized as a weight to cast appropriate loss punishment in different branches during the optimization of the proposed aggregated landmark detection as follows:  

\begin{equation}
\mathcal{L}_T = \dfrac{1}{LK}\sum\limits_{i=1}^L\sum\limits_{k=1}^K Sc_{i,k} ||p^*_{i} - \tilde{p}_{i,k}||_2\; , 
\label{aggreg}
\end{equation}

\noindent where $\tilde{p}_{i,k}$ represents the $i$-th estimated landmark at $k$-th branch and $p^*_{i}$ indicates the ground truth for $i$-th landmark location.

Given a test image, we extract the rough estimation of the landmark coordinates employing the trained landmark detector $\Phi$ in the aggregated approach and consider them as the coarse landmarks $P$. Afterward, we perform the manipulation approach on the extracted landmarks $P$ and generate manipulated images. The extracted landmarks and manipulated images are used in the aggregated framework to produce the final landmarks. The final landmarks are calculated as follows: 

\begin{equation}
p_i^{f}  = \sum\limits_{k=1}^K Sc_{i,k}. (\tilde{p}_{i,k}) \; , 
\label{final_testing}
\end{equation}
\noindent where $p_i^f$ is the coordinate of the $i$-th landmark employing the proposed aggregated network such that $\Phi (I) = P^f$ for $L$ landmark locations. 

As it is mentioned in the manuscript, during the training phase the manipulation is performed on the coarse landmarks' locations and we have access to them. However, given a test image, we extract the landmarks using the trained landmark detector $\Phi$ and then use them as the coarse landmarks' locations in the aggregated framework to predict the final landmark locations.

\begin{figure*}[t]
\begin{center}

\includegraphics[width=.8\linewidth]{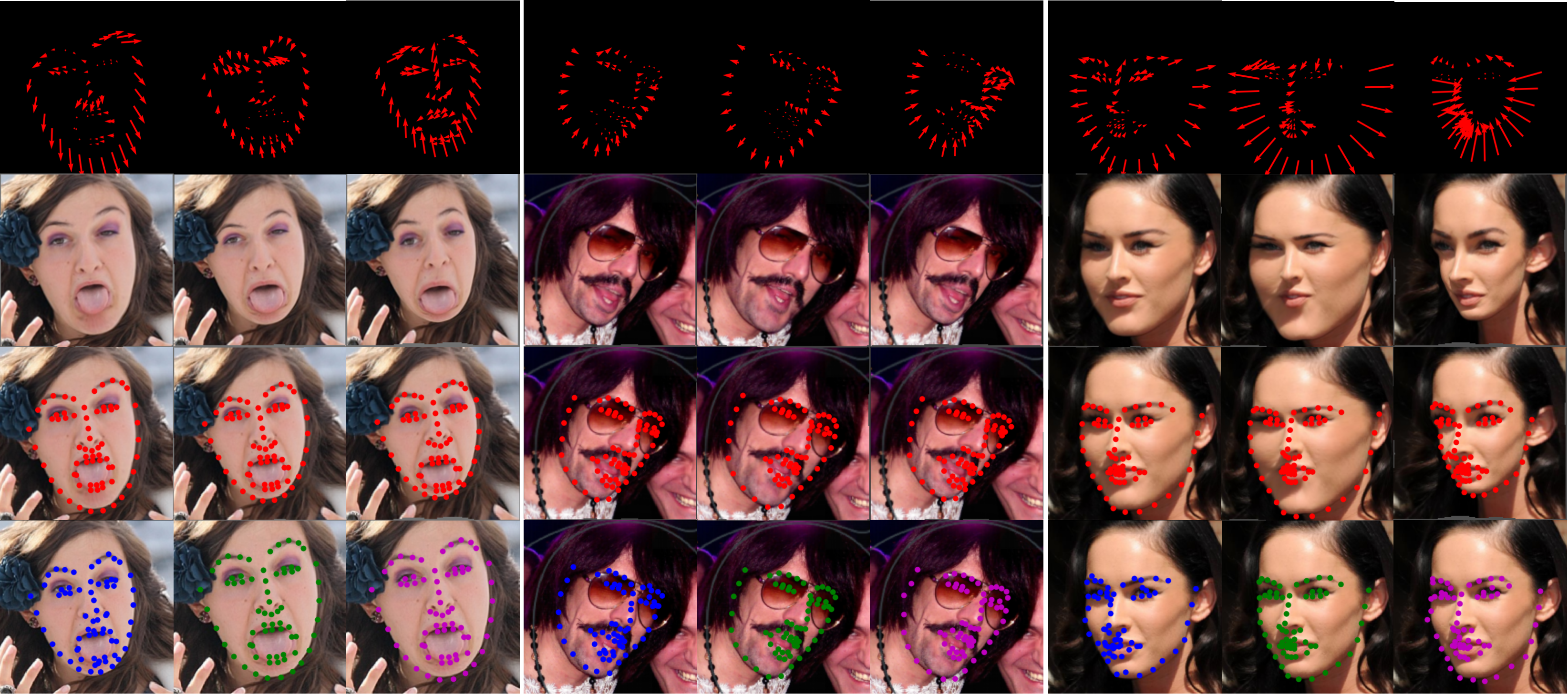}
\vspace{-15pt}
\end{center}
   \caption{The representative results for three face images from the 300-W dataset. For each face, the first row represents displacement fields for the aggregated network with $K=3$ (the arrows are exaggerated for the sake of illustration). The second row shows manipulated images using the corresponding displacement field, and the third row represents the extracted landmarks given the corresponding manipulated images to the landmark detector,$ \Phi(\hat{I})$. The fourth row represents landmarks' locations on the input image $I$ from the base detector (in blue), ground-truth (in green), and GEAN landmark detector (in magenta), respectively.}
\label{fig:long}
\label{fig:onecol}
\end{figure*}

One question that comes to mind is: what if the predicted landmark locations using the trained landmark detector $\Phi$ are not an accurate representation for the original coarse landmarks? To compensate this issue and make the conditions equal for the training and testing phases, we add random noise to the ground truth landmarks such that $P = P^*+ \eta$ where $P^*$ is the ground truth for landmark coordinates and $\eta$ is random noise. Afterward, we employ these landmarks as the coarse landmarks $P$ in the aggregated framework during the training phase. 


\begin{table*}[]
\centering 
\small
\setlength\tabcolsep{1pt} 
\begin{tabular}{c|cccccccc|ccc}\hline
Methods    & \rotatebox{30}{ERT~\cite{6909637}}  & \rotatebox{30}{LBF~\cite{ren2016face}}  & \rotatebox{30}{CFSS~\cite{7299134}} & \rotatebox{30}{CCL~\cite{7780740}}  & \rotatebox{30}{Two-St.~\cite{8099876}} & \rotatebox{30}{SAN~\cite{dong2018style}}  & \rotatebox{30}{ODN~\cite{zhu2019robust}}&\rotatebox{30}{LRef.~\cite{su2019efficient}}&  \rotatebox{30}{GEAN$_{adv}$}  & \rotatebox{30}{GEAN$_{GK}$}&\rotatebox{30}{GEAN$_{Gadv}$} \\
\hhline{============}
AFLW-Full  & 4.35 & 4.25 & 3.92 & 2.72 & 2.17      & 1.91 &1.63& 1.63& 1.69 &1.64&\textbf{1.59} \\ \hline
AFLW-Front & 2.75 & 2.74 & 2.68 & 2.17 & -         & 1.85 &1.38& 1.46&1.44 &1.38 &\textbf{1.34}\\\hline
\end{tabular}
\caption{Comparison of different methods based on normalized mean errors (NME) on AFLW dataset.}
\label{TableALFW}
\vspace{-20pt}
\end{table*}

\section{Experiments}
In the following section, we consider three variations of our GEAN approach. $GEAN_{adv}$~(\ref{Landmarkadv}) represents the case when the manipulated faces are generated using the adversarial attack approach. $GEAN_{Gadv}$ (\ref{Landmark_Groupadv}) and $GEAN_{GK}$ (\ref{Landmark_Groupknown}) represent the cases when the manipulated faces are generated using the semantically grouped adversarially attack and known transformations approach, respectively. In order to show the effectiveness of GEAN we evaluate its performance on three following datasets:

\textbf{300-W~\cite {sagonas2013300}:} The dataset annotates five existing datasets with 68 landmarks: LFPW~\cite{belhumeur2013localizing}, AFW~\cite{6248014}, HELEN~\cite{le2012interactive}, iBug, and XM2VTS. Following the common setting in~\cite{dong2018style,8099876}, we consider 3,148 training images from LFPW, HELEN, and the full set of AFW. The testing dataset is split into three categories of common, challenging, and full groups. The common group contains 554 testing images from LFPW and HELEN datasets, and the challenging test set contains 135 images from the IBUG dataset. Combining these two subsets form the full testing set.      

\textbf{AFLW~\cite{6130513}:} This dataset contains 21,997 real-world images with 25,993 faces in total with a large variety in appearance (\textit{e.g.}, pose, expression, ethnicity, and age) and environmental conditions. This dataset provides at most 21 landmarks for each face. Having faces with different pose, expression, and occlusion makes this dataset challenging to train a robust detector. Following the same setting as in~\cite{8099876,dong2018style}, we do not consider the landmark of two ears. This dataset has two different categories of AFLW-Full and AFLW-Frontal~\cite{7780740}. AFLW-Full contains 20,000 training samples and 4,386 testing samples. AFLW-Front uses the same set of training samples as in AFLW-Full, but only contains 1,165 samples with the frontal face for the testing set.

\textbf{COFW~\cite{burgos2013robust}:} This dataset contains 1,345 images for training and 507 images for test. Originally this dataset annotated with 21 landmarks for each face. However, there is a new version of annotation for this dataset with 68 landmarks for each face~\cite{ghiasi2015occlusion}. We used a new version of annotation to evaluate proposed method and comparison with the other methods.

\textbf {Evaluation:} Normalized mean error (NME) and and Cumulative Error Distribution (CED) curve are usually used as metric to  evaluate performance of different methods~\cite{7780740,8099876}. Following~\cite{ren2016face}, we use the inter-ocular distance to normalize mean error on 300-W dataset. For the AFLW dataset we employ the face size to normalize mean error as there are many faces with inter-ocular distance closing to zero in this dataset~\cite{8099876}.


\textbf {Implementation Details:} 
We employ the face recognition model developed by Schroff et al.~\cite{schroff2015facenet} which obtain the state-of-the-art accuracy on the Labeled Faces in the Wild (LFW)~\cite{huang2008labeled} dataset as the face recognizer. We train this model on more than 3.3M training images and the average of 360 images per ID (subject) from VGGFace2 dataset~\cite{cao2018vggface2} to recognize 9,101 celebrities. The landmarks are divided to five different categories based on facial regions as: 1) $P_1:$ right eye and eyebrow, 2) $P_2:$ left eye and eyebrow, 3) $P_3:$ nose, 4) $P_4:$ mouth, and 5) $P_5:$ jaw. The number of landmarks in each group is as: \{$n_1=11$, $n_2=11$, $n_3=9$, $n_4=20$, $n_5=17$\}. We set $\tau=0.6$, $\delta$ to $5\%$ of the width of the bounding box of each face.

The landmarks' coordinates are scaled to lie inside the range $[-1,1]^2$ where $(-1,-1)$ is the top left corner and $(1,1)$ is the bottom right corner of the face image. All the coordinates are assumed to be continuous values since TPS has no restriction on the continuity of the coordinates because of the differentiable bilinear interpolation~\cite{NIPS2015_5854}. The face images are cropped and resized to $(256 \times 256)$. We follow the same setting in~\cite{8014987} and use four stacks of hourglass network for the landmark detection network. We train our model with the batch size of 8, weight decay of $5\times 10^{-4}$, and the starting learning rate of $5\times 10^{-5}$ on two GPUs. The face bounding boxes are expanded by the ratio of 0.2 and random cropping is performed as data augmentation.    
\begin{table}[t]
\centering
\small
\setlength\tabcolsep{2pt}
\begin{tabular}{c|c|c|c} \hline
Method           & Common & Challenging & Full Set \\ \hhline{====}
LBF~\cite{ren2016face}              & 4.95   & 11.98       & 6.32     \\
CFSS~\cite{7299134}             & 4.73   & 9.98        & 5.76     \\
MDM~\cite{Trigeorgis2016MnemonicDM}              & 4.83   & 10.14       & 5.88     \\
TCDCN~\cite{Multitask}            & 4.80   & 8.60        & 5.54     \\
Two-Stage~\cite{8099876} & 4.36   & 7.42        & 4.96     \\
RDR~\cite{8237443}              & 5.03   & 8.95        & 5.80     \\
Pose-Invariant~\cite{Jourabloo2017PoseInvariantFA}   & 5.43   & 9.88        & 6.30     \\
SAN~\cite{dong2018style}      & 3.34   & 6.60        & 3.98     \\ 
ODN~\cite{zhu2019robust}      & 3.56   & 6.67        & 4.17     \\
LRefNets~\cite{su2019efficient}      & 2.71   & 4.78        &3.12     \\ \hline
\textbf{GEAN}     & \textbf{2.68}   & \textbf{4.71}        & \textbf{3.05}    \\\hline
\end{tabular}

\caption{Normalized mean errors (NME) on 300-W dataset.}
\label{Table300W}
\vspace{-15pt}
\end{table}

\subsection{Comparison with State-of-the-arts Methods:}\label{stateofart}
\textbf {Results on 300-W.} Table~\ref{Table300W} shows the performance of different facial landmark detection methods on 300-W dataset. We compare our method to the most recent state-of-the-art approaches in the literature~\cite{dong2018style,8237443,Jourabloo2017PoseInvariantFA,su2019efficient}. The number of branches in training and testing phases is set to $K=5$. Among the three proposed approaches, we consider $GEAN_{Gadv}$ as the final proposed method to compare with the state-of-the-art methods.  The results show its superiority compared to the other methods for both types of bounding boxes. 
The superiority of the proposed method shows the effect of manipulated images which target the important locations in the input face image. Aggregation of these images improves the facial landmark detection by giving more attention to the keypoint locations of the face images.  

\textbf {Results on AFLW.} We conduct our experiments on the training/testing splits  and the bounding box provided from~\cite{7780740,7299134}.  Table~\ref{TableALFW} shows the effectiveness of proposed GEAN. AFLW dataset provides a comprehensive set of unconstrained images. This dataset contains challenging images with rich facial expression and poses up to $\pm120^{\circ}$ for yaw and $\pm90^{\circ}$ for pitch and roll. Evaluation of proposed method on this challenging dataset shows its robustness to large pose variations. Indeed, the weighted aggregation of predictions obtained on the set of deformed faces reduces the sensitivity of GEAN to large pose variations.


\begin{table*}[]
\centering 
\small
\begin{tabular}{ccccc|cccc|ccccc}\hline
    & \multicolumn{4}{c}{\textbf{Common test set}} & \multicolumn{4}{c}{\textbf{Challenging test set}} & \multicolumn{5}{c}{\textbf{Full test set}} \\ \cline{2-14} 
\multicolumn{1}{c}{\textbf{Train}} & 1    & 3               & 5               & 7 & 1    & 3               & 5               & 7& 1$^{*}$ & 1    & 3               & 5               & 7      \\ \hhline{==============}
\multicolumn{1}{c|}{1}              & 4.40 & 3.77            & 3.48            & 3.43  & 5.44 & 5.35            & 5.30            & 5.28 & 4.80 & 4.80 & 4.49            & 4.07            & 4.02     \\
\multicolumn{1}{c|}{3}              & 3.67 & 3.25            & 3.03            & 2.98  & 5.33 & 5.27            & 5.18            & 5.10 & 4.68 & 4.46 &    4.01        &  3.77           &  3.74    \\
\multicolumn{1}{c|}{5}              & 3.35 & 2.99            & 2.68            & 2.66 & 5.26 & 4.97            & 4.71            & 4.67    & 4.63 & 4.04 & 3.64            & 3.05            & 3.01  \\
\multicolumn{1}{c|}{7}              & 3.32 & 2.93            & 2.65            & 2.63  & 5.22 & 4.90            & 4.65             & 4.60 & 4.59 & 3.96 &  3.56          & 3.00            & 2.97   \\\hline
\end{tabular}

\caption{Comparison of NME on three test sets of 300-W with different numbers of branches for the training and testing. The column with asterisk demonstrates the results for evaluating the performance of our model without aggregation.}
\label{ablation1}
\vspace{-15pt}
\end{table*}

\begin{figure}[t]
\begin{center}
 \includegraphics[width=\linewidth]{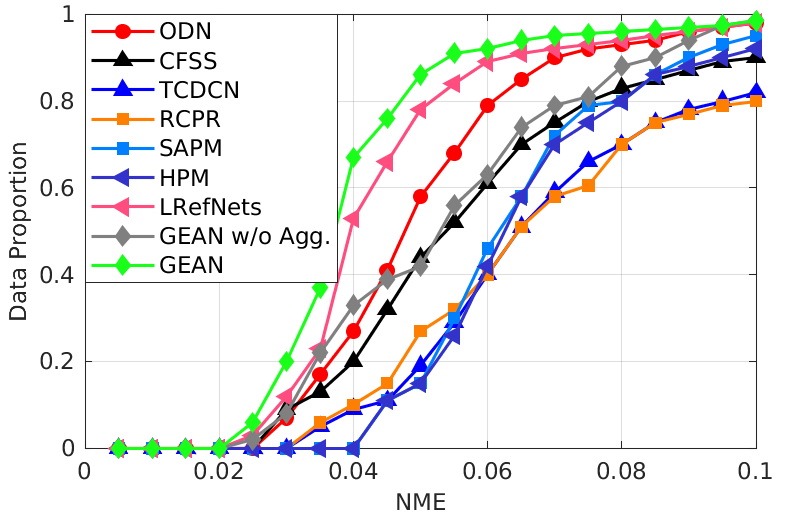}
   
\end{center}
\caption{Comparison results of different methods (ODN~\cite{zhu2019robust}, CFSS~\cite{7299134}, TCDCN~\cite{zhang2015learning}, RCPR~\cite{burgos2013robust}, SAPM~\cite{ghiasi2015using}, HPM~\cite{ghiasi2014occlusion}, LRefNets~\cite{su2019efficient}, and GEAN) on COFW dataset.}
\label{fig:NME-CED}
\vspace{-10pt}
\end{figure}

\textbf{Results on COFW.} Figure~\ref{fig:NME-CED} shows the evaluation of our proposed method in a cross-dataset scenario. We conduct evaluation using the models trained on the 300-W dataset and test them on re-annotated COFW dataset with 68 landmarks~\cite{ghiasi2015occlusion}. The comparison is performed using the CED curves as plotted in Figure~\ref{fig:NME-CED}. The best performance belongs to our method (GEAN) with 4.24\% mean error compared to the previous best~\cite{su2019efficient} with 4.40\% mean error. This shows the robustness of our method compared to other state-of-the-art methods in detecting facial landmarks. 

Timing of the proposed approach directly depends on the number of branches and also the approach that we take to generate the manipulated faces. It is shown in~\cite{Dabouei2019FastGA} that semantically grouping the landmark locations increases the time of manipulated faces generation. However, it can overcome the problem of face distortion due to considering the semantic grouping. Therefore, there is a trade-off between the speed and accuracy of the proposed framework. However, in the case of aggregating with five branches and employing $GEAN_{Gadv}$ for generating manipulated faces, the framework runs in 17 FPS with NVIDIA TITAN X GPU.

\subsection{Ablation Studies}

\textbf{Number of branches:} In this section, we observe the effect of adding branches on the performance of the aggregated framework. We start with $k=1$ in which there is no aggregation and one manipulated image is generated. We increase the number of branches from one to seven and measure the performance of aggregated network on the common, challenging, and full split of the 300-W dataset.


In addition, the number of branches in the training and testing phases is not necessarily the same. For example, the number of branches in the aggregated framework can be three while the number of branches in the testing phase is equal to 10. This is essentially important due to the time complexity of the framework during the training and testing phases. In addition, one can train the network on two or three branches while test it on more branches to get more accurate results. Table~\ref{ablation1} shows the evaluation results of 16 training and testing combinations, \textit{i.e.}, four different training architectures ($K=1,3,5,7$) multiply four different testing architectures on 300-W common, challenging, and full test set, respectively. 

As we can observe, the performance will be increased if the number of branches is increased during the training phase. However, we observe that adding more than five branches to the framework does not significantly improve the results with the cost of more computational complexity. The same behavior is observed for the testing framework. By increasing the number of branches in the testing phase, the accuracy is increased. This is useful when we want to reduce the computational complexity in training and maintaining the performance in the testing phase to some extent. Considering both accuracy and speed, we choose the framework with the number of training and testing branches equal to five for the sake of comparison with state-of-the-art (\ref{stateofart}).

We also conduct another experiment to demystify the effect of aggregation part in the proposed GEAN. In this case, GEAN with just one branch is trained on all the deformed and manipulated faces without the aggregation part. Table~\ref{ablation1} and Figure~\ref{fig:NME-CED} show the performance of $GEAN~w/o~Agg.$ compared to the proposed GEAN. For the sake of fair comparison, we trained the network on the same number of manipulated face images for both methods. By comparing column (1$^{*}$) with column (1) of 300-W full test set, it is shown that the proposed GEAN which is trained with the exact same faces is superior to its counterpart without aggregation. Figure~\ref{fig:NME-CED} also confirms the effectiveness of aggregation part and illustrates the fact that proposed GEAN performs beyond a careful augmentation.   


\textbf{A Comparison between Three Different Variations of GEAN:} Three different approaches of $GEAN_{adv}$, $GEAN_{Gadv}$, and $GEAN_{GK}$ have been introduced in this paper. Through this section, we evaluate the performance of three different variations of our GEAN method on AFLW dataset. As Table~\ref{TableALFW} shows, both $GEAN_{Gadv}$ and $GEAN_{GK}$ outperform $GEAN_{adv}$ approach. We attribute this to the fact that $GEAN_{adv}$ does not consider grouping different landmarks semantically. This causes inconsistent displacements for the landmarks of one region (\textit{e.g.}, left eye) and generate distorted images. In addition, the amount of displacement of landmarks in manipulated images might be greater than the manipulated images with the other two methods. However, this displacement might not be beneficial as it does not consider the general shape of each face region. Utilizing clipping constraint can mitigate this issue to some extent. However, this approach still suffers from not considering the semantic groups. 

$GEAN_{Gadv}$ works the best among all three proposed approaches. Several reasons can explain this superiority. This approach considers the semantic relationship among the landmarks of same regions of the face. In addition, the manipulated images in this approach have different face IDs from the original face image. Therefore, in the framework with $K$ branches, the aggregation is performed in $K$ different face IDs. This makes this approach to preserve a reasonable relative distance among different groups of landmarks since it could fool the recognizer to misclassify it. However, this is not necessarily the case for the $GEAN_{GK}$ approach. This makes the $GEAN_{Gadv}$ to capture more important landmark displacement for the image manipulation which is beneficial for the aggregation. The advantages of the other two approaches (\textit{i.e.}, adversarially attack technique in $GEAN_{adv}$ and semantic grouping of landmarks in $GEAN_{GK}$) is unified in $GEAN_{Gadv}$ which leads to a better landmark detection performance. 

\section{Conclusion}
In this paper, we introduce a novel approach for facial landmark detection. The proposed method is an aggregated framework in which each branch of the framework contains a manipulated face. Three different approaches are employed to generate the manipulated faces and two of them perform the manipulation via the adversarial attacks to fool a face recognizer. This step can decouple from our framework and potentially used to enhance other landmark detectors~\cite {dong2018style, 8099876,8014987}. Aggregation of the manipulated faces in different branches of GEAN leads to robust landmark detection. An ablation study is performed on the number of branches in training and testing phases and also on the effect different approaches of face image manipulation on the facial landmark detection. The results on the AFLW, 300-W, and COFW datasets show the superiority of our method compared to the state-of-the-art algorithms.   

{\small
\bibliographystyle{ieee}
\bibliography{egbib}
}

\end{document}